\documentclass[runningheads]{llncs}

\usepackage{eccv}

\usepackage{epsfig}
\usepackage{amsmath}
\usepackage{amssymb}
\usepackage{tabularx} 
\usepackage{ragged2e} 
\usepackage{booktabs} 
\usepackage{overpic}
\usepackage{bm}
\usepackage{bbding}
\usepackage{multirow}

\usepackage{eccvabbrv}

\usepackage{graphicx}
\usepackage{booktabs}

\usepackage[accsupp]{axessibility}  

\usepackage{hyperref}

\usepackage{orcidlink}

\begin{document}

\title{WildRefer: 3D Object Localization in Large-scale Dynamic Scenes with Multi-modal Visual Data and Natural Language} 

\titlerunning{WildRefer}

\author{Zhenxiang Lin\inst{1} 
\and
Xidong Peng\inst{1} 
\and
Peishan Cong\inst{1}
\and
Ge Zheng\inst{1}
\and
Yujin Sun\inst{2}
\and
Yuenan Hou\inst{3}
\and
Xinge Zhu\inst{4}
\and
Sibei Yang\inst{1}
\and
Yuexin Ma\inst{1}\thanks{Corresponding author. This work was supported by NSFC (No.62206173), Natural Science Foundation of Shanghai (No.22dz1201900), Shanghai Sailing Program (No.22YF1428700), MoE Key Laboratory of Intelligent Perception and Human-Machine Collaboration (ShanghaiTech University), Shanghai Frontiers Science Center of Human-centered Artificial Intelligence (ShangHAI).}
}

\authorrunning{Lin et al.}

\institute{
$^1$ShanghaiTech University 
$^2$The University of Hong Kong \\
$^3$Shanghai AI Laboratory 
$^4$The Chinese University of Hong Kong \\
\email{zhenxianglin94@gmail.com}, 
\email{mayuexin@shanghaitech.edu.cn}
}

\makeatletter
\let\@oldmaketitle\@maketitle
\renewcommand{\@maketitle}{
   \@oldmaketitle
 \begin{center}
      \includegraphics[width=1.0\linewidth]{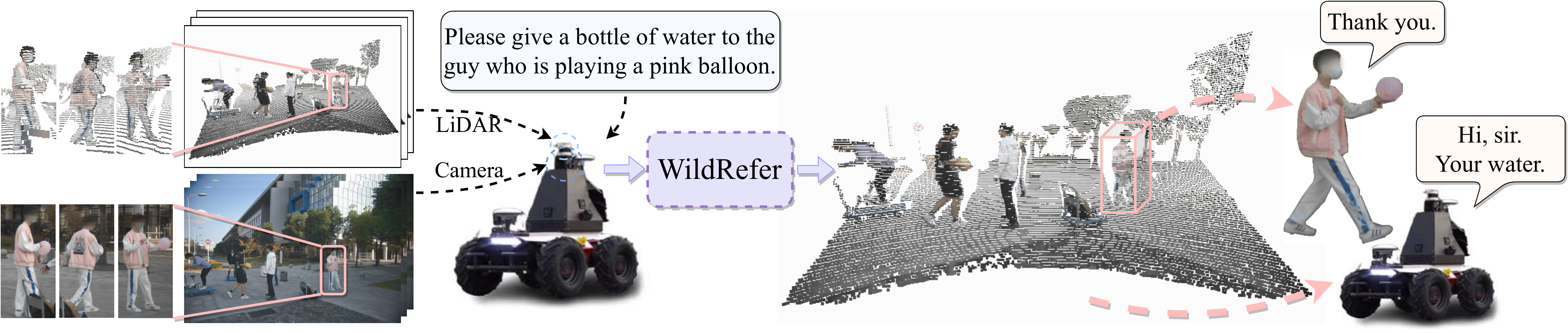}
 \end{center}
  \refstepcounter{figure}\normalfont Figure~\thefigure. 
  Introduction to the 3DVGW task and related application. The assistive robot observes the dynamic scene and locates the 3D object in the physical world according to natural language descriptions. Then, the robot moves to the target object to provide service. WildRefer provides a LiDAR-camera multi-sensor-based solution, which can conduct 3D visual grounding in large-scale unconstrained environment.
  \label{fig:teaser}
  \newline
  }
\makeatother

\maketitle

\begin{abstract}
   We introduce the task of 3D visual grounding in large-scale dynamic scenes based on natural linguistic descriptions and online captured multi-modal visual data, including 2D images and 3D LiDAR point clouds. We present a novel method, dubbed WildRefer, for this task by fully utilizing the rich appearance information in images, the position and geometric clues in point cloud as well as the semantic knowledge of language descriptions. Besides, we propose two novel datasets, \emph{i.e.}, STRefer and LifeRefer, which focus on large-scale human-centric daily-life scenarios accompanied with abundant 3D object and natural language annotations. Our datasets are significant for the research of 3D visual grounding in the wild and has huge potential to boost the development of autonomous driving and service robots. Extensive experiments and ablation studies demonstrate that our method achieves state-of-the-art performance on the proposed benchmarks. The code is provided in \url{https://github.com/4DVLab/WildRefer}.
\end{abstract}


\section{Introduction}
\label{sec:introduction}

The human capacity for real-time scene analysis, dynamic object localization, and semantic comprehension empowers us to perform various tasks in real life, such as assisting an elderly person who has fallen or serving tea to a lady engaged in a dance. In this paper, we introduce a new task, \textbf{3D Visual Grounding in the Wild (3DVGW)}, which aims at locating a specific object within the physical environment based on online captured visual data and linguistic descriptions, as depicted in Figure.~\ref{fig:teaser}. This task holds substantial importance for many downstream applications, such as assistive robots, autonomous driving, surveillance, \emph{etc}.

In recent years, significant strides have been made in the field of 2D visual grounding~\cite{mattnet, dga, vilbert, resc, transvg}, largely driven by the availability of numerous large-scale datasets~\cite{refcoco, refcocog, clevr, visualgenome, guesswhat}. However, these methods and datasets are primarily confined to the realm of 2D domains, making their direct applicability to 3D scenarios a challenging proposition. To effectively capture the 3D geometric attributes of objects and spatial relationships among them for 3D visual grounding, various endeavors~\cite{scanrefer,referit3d,instancerefer,3dvgtransformer,3dvgn,3djcg,3dsps,butd-detr, eda} have introduced related datasets and solutions, which are meaningful for the robotic manipulation and augmented reality. Nevertheless, RGB-D cameras are typically limited to indoor environments with a sensing range of up to 5 meters, and these methods are designed for pre-scanned indoor static scenarios. Consequently, they present certain limitations when applied to the dynamic, large-scale scenes with online visual data.

To capture long-range scenes in free environment, we leverage calibrated and synchronized LiDAR and camera as visual sensors, which are usually adopted by outdoor robots and autonomous vehicles~\cite{nuscenes,Semantickitti,zhu2021cylindrical}. 3DVGW differs from previous RGB-D-based visual grounding in two aspects, including the dynamic scenario and the multi-modal visual data. For the dynamic real world, motion-related language descriptions are commonly used and temporal information in visual data is worth exploring for more accurate vision-language matching. For multi-modal visual data, LiDAR point cloud contains precise 3D geometric and position information and images embrace rich appearance clues, which calls for an effective and efficient multi-modal fusion strategy to make full use of the complementary modalities and further enhance the visual grounding accuracy.

Taking the aforementioned issues into consideration, we propose a novel approach, namely \textbf{WildRefer}, for 3DVGW based on multi-modal signals. First, we design a \textbf{Dynamic Visual Encoder} to extract the motion feature from consecutive visual data. Considering that the location, pose, and appearance feature of objects may vary over time, it is nontrivial to track the changes belonging to the same object. We apply a transformer-based mechanism to match corresponding sequential features automatically. Second, we design a \textbf{Triple-modal Feature Interaction} module to make the semantic information in language, the geometry information in point clouds, and the appearance information in images interact with each other to enhance the representation of both linguistic and visual features. Finally, we adopt a DETR-form decoder to predict bounding boxes of objects and conduct the selection for target through contextualized features of language utterance and visual tokens.

To facilitate the research of 3DVGW, we propose two new datasets, STRefer and LifeRefer, which are captured in large-scale human-centric scenarios by a 128-beam mechanical LiDAR and an industrial camera. STRefer is extended from the open-source dataset STCrowd~\cite{stcrowd}, which targets for 3D perception task in crowd scenarios. We sample 662 LiDAR scans and synchronized images and add 5,458 free-form language descriptions. LifeRefer is extended from the huge dataset HuCenLife~\cite{hucenlife}, collected in various daily-life scenes with rich human activities, human-human interactions, and human-object interactions. We annotate 25,380 natural language descriptions for 11,864 subjects from 3,172 scenes. Human-centric free environment is much more complex than static indoor scenes or traffic scenes, which is challenging but significant for the development of autonomous driving and service robots. Our datasets fill the blank of data in this field and have potential to boost related research.
In particular, our method is evaluated on both datasets and achieves state-of-the-art performance.

Our contributions can be summarized as follows:

\begin{itemize}
    \item We introduce the task of 3D vision grounding in the wild, where the scene is large-scale and dynamic, and the visual data is online captured by multiple sensors.
    \item We provide two novel datasets for 3DVGW, which are captured in human-centric scenarios, containing multi-modal visual data and free-form natural language descriptions.
    \item  We propose a novel method for 3DVGW, where temporal information is fully explored and multi-modal features are fused effectively for accurate target localization, which achieves state-of-the-art performance on both STRefer and LifeRefer.
    
\end{itemize}
\section{Related Works}
\label{sec:related_works}

\subsection{2D Visual Grounding}
There are already a lot of investigations about image-based or video-based 2D visual grounding task. To handle this task, a number of datasets~\cite{refcoco,refcocog,visualgenome,guesswhat,zhou2019grounded,rohrbach2017generating} are created by asking the annotators to provide expressions for specific image regions, and many methods~\cite{cho2021unifying,huang2021deconfounded,mdetr,zhu2022seqtr} are proposed subsequently. Some approaches are based on the two-stage architecture~\cite{mil,mattnet,dga,cmn,lgran} by detecting all potential objects from the image, and then locating the target object according to the referring expression. Inspired by traditional one-stage detection methods, one-stage grounding approaches~\cite{resc, realtimemodel, fastonestage} appear by fusing image and language features densely and locating the target directly. With the success of transformer~\cite{transformer} and BERT~\cite{bert}, recently, some researchers~\cite{vilbert,vl-bert,transvg} apply these frameworks on the visual grounding problem and achieve promising performance. However, due to the lack of depth and 3D geometry information in images, these 2D visual grounding approaches cannot be extended to 3D scenes directly. 

\subsection{3D Visual Grounding}
To facilitate the development of artificial intelligence in 3D real scenes, 3D visual grounding has gained more and more attention. ScanRefer~\cite{scanrefer} is the first indoor 3D grounding dataset based on RGB-D scans and natural language descriptions. Afterwards, Achlioptas et al.\cite{referit3d} propose Nr3D and Sr3D datasets according to different language forms. Moreover, Zhu et al.\cite{zhu20233d} combine current 3DVG datasets and collect a new dataset to increase the volume of data for pretraining. These datasets stimulate the progress of visual grounding in 3D scenes. Yuan et al.\cite{instancerefer} use instance-level information to enhance the location. Huang et al.\cite{mvt} use multiple views to improve the robustness. Zhao et al.\cite{3dvgtransformer} propose a new framework which takes more attention on spatial relation among point clouds. Cai et al.\cite{3djcg} design a unified framework to solve both visual captioning and grounding tasks. Feng and Li\cite{3dvgn} apply GCN to improve the correlations between object and object as well as object and scene. Liu et al.\cite{liu2021refer} propose a bottom-up approach to gradually aggregate content-aware information for grounding the referred object. Chen et al.\cite{ham} develop a hierarchical attention model and improve the performance. Jain et al.\cite{butd-detr} use a top-down method to compensate the missing objects from detection issue. Based on it, Wu et al.\cite{eda} decouple the language and achieves a higher performance. Luo et al.\cite{3dsps} directly locate the target by a one-stage method. However, these 3D visual grounding datasets and methods are all based on RGB-D scans in static indoor scenes. Furthermore, similar with popular 2D vision-language methods, Chen et al.\cite{unit3d} apply the pretrained unique transformer to address downstream 3D vision-language problem.
M3DRefer-CLIP \cite{m3drefer} uses pretrained CLIP as image and text encoder to improve the performance. 
Another dataset KITTI360Pose~\cite{text2pos} focuses on outdoor traffic scenes, but it targets for the mobile robot localization with templated language descriptions for only positions. To our knowledge, our datasets and methods are the first for 3D visual grounding task in large-scale dynamic daily-life scenes based on LiDAR and camera multiple visual sensors.

\subsection{LiDAR-camera-fusion based 3D Detection}
To better utilize the advantages of LiDAR and camera, more and more sensor-fusion works~\cite{qi2018frustum,qi2020imvotenet,sindagi2019mvx,pointpainting,XuYan20222DPASS2P, Zhuang_2021_ICCV, liang2022bevfusion} appear, which can be roughly divided into data-level and feature-level fusion strategies. The former usually attaches the image information to points according to the calibration matrix to enhance the point cloud feature. It misses lots of image features of the dense representation. The latter usually extracts individual visual features and fuses high-level features directly by concatenation or multiplication, but lacks interpretable correspondence between two modalities. Recently, transformer-based fusion methods~\cite{bai2022transfusion,han2022licamgait,li2022deepfusion,chen2022futr3d, cong2022weakly} achieve impressive performance by making the network learn the projection and fuse valuable features from different modalities. However, they all target for vision-based perception tasks without considering the interaction with language features. We propose an effective fusion strategy for triple-modal features from distinct visual and linguistic inputs.
\section{Dataset}
\label{sec:dataset}

\begin{table*}
\centering
\begin{center}
\caption{Comparison of 3D visual grounding datasets, where ``bbx'' indicates 3D bounding box. For the scene range, 30m is the approximated perception radius with annotations for our LiDAR sensor, while 10m is the range of the whole scene pre-scaned by RGB-D camera. For KITTI360Pose, the sequential point clouds are stitched together for annotation, so the number of scenes and range are unable to count.}
\label{table:dataset_comparison}
\resizebox{\linewidth}{!}{
\begin{tabular}{c|c|c|c|c|c|c|c|c|c|c}
\hline
Dataset & Object &Target & Data    & Language & 3D  & \multicolumn{4}{c|}{Scene} & Time    \\ \cline{7-10} 
 &     Number &    &   Modality         & Form &  Label        & Range & Number & Indoor & Outdoor &Series\\
\hline
Sr3d~\cite{referit3d}  & 8,863 &furniture  & RGB-D         & templated &  bbx & 10m & 1,273 & \multicolumn{1}{c|}{\textcolor{green}{\Checkmark}} & \textcolor{red}{\XSolidBrush}  & \textcolor{red}{\XSolidBrush}  \\
Nr3d~\cite{referit3d}  & 5,878  &furniture & RGB-D          & free &  bbx& 10m & 641 & \multicolumn{1}{c|}{\textcolor{green}{\Checkmark}} & \textcolor{red}{\XSolidBrush} & \textcolor{red}{\XSolidBrush}  \\
ScanRefer~\cite{scanrefer}  & 11,046 &furniture & RGB-D        & free &  bbx & 10m & 800 & \multicolumn{1}{c|}{\textcolor{green}{\Checkmark}} & \textcolor{red}{\XSolidBrush} & \textcolor{red}{\XSolidBrush}  \\
ScanScribe~\cite{zhu20233d}  & 56,100 & variety & RGB-D        & templated \& free  &  bbx & 10m &  2,995 & \multicolumn{1}{c|}{\textcolor{green}{\Checkmark}} & \textcolor{red}{\XSolidBrush} & \textcolor{red}{\XSolidBrush}  \\
 KITTI360Pose~\cite{text2pos}  & 14,934  &location & image\& LiDAR point cloud      & templated &  position& -- & -- & \textcolor{red}{\XSolidBrush}  & \textcolor{green}{\Checkmark} & \textcolor{red}{\XSolidBrush}   \\
\hline 
\textbf{STRefer} & 3,581 &human & image\& LiDAR point cloud & free &  bbx& 30m &662& \multicolumn{1}{c|}{\textcolor{red}{\XSolidBrush}} & \textcolor{green}{\Checkmark} & \textcolor{green}{\Checkmark}  \\
\textbf{LifeRefer} & 11,864 &human  & image\& LiDAR point cloud & free&  bbx & 30m& 3,172& \multicolumn{1}{c|}{\textcolor{green}{\Checkmark}} & \textcolor{green}{\Checkmark} & \textcolor{green}{\Checkmark} \\
\hline
\end{tabular}
} 
\end{center}
\end{table*}

As shown in Table.~\ref{table:dataset_comparison}, previous 3D visual grounding datasets~\cite{referit3d,scanrefer} usually focus on pre-scanned static indoor scenes. Although KITTI360Pose~\cite{text2pos} shifts the focus to the outdoor traffic scenario, it targets for mobile robot localization in city-level scenes, which are still static without time series and without 3D labels for objects. In order to benefit the development of service robots, we capture and annotate large-scale data for 3D visual grounding from daily human-centric scenarios with rich human actions, human-human interactions, and human-object interactions. We introduce detailed data acquisition, annotation procedure, statistics, and characteristics of our datasets in the following.

\subsection{Data Acquisition}
We contribute two datasets for the research of 3DVGW, including STRefer and LifeRefer. \textbf{STRefer} is built on an open-source dataset STCrowd~\cite{stcrowd}, which mainly focuses on 3D perception tasks in dense crowd scenarios. \textbf{LifeRefer} extends the scenes from open-source dataset HuCenLife~\cite{hucenlife}, which is collected in diverse large-scale scenes with multi-person activities, including shopping mall, gym, playground, park, campus, etc. Such scenarios are complex with abundant human motions and interactions with the environment, which provides rich data but also bring challenges for accurate 3D localization. For both datasets, the visual data are captured by calibrated and synchronized 10Hz 128-beam Ouster-OS1 LiDAR and industrial camera with $1920\times1200$ image resolution. For all annotated humans in these datasets, we add free language descriptions.

\begin{figure*}
\centering
\includegraphics[width=1.0\textwidth]{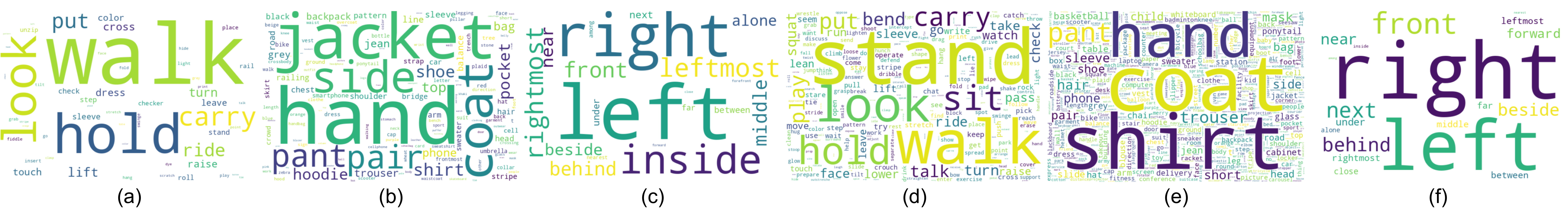} 
\caption{(a), (b), (c) and (d), (e), (f) are word clouds for action words, non-human object words, and spatial relation words on STRefer and LifeRefer, respectively.}
\label{fig:word_cloud}
\end{figure*}
\subsection{Annotations}

For each target object, we provide a free-form linguistic description and a 3D bounding box. The 3D bounding box is represented by $(x, y, z, l, w, h, \theta)$, where $x, y, z$ denotes the center in the LiDAR coordinates system, $l, w, h$ represents the size of box, and $\theta$ is the orientation. The interval of two consecutive frames is 0.4s for STRefer and 1s for LifeRefer. To accurately describe the human motion, annotators are allowed to see previous three frames when labeling current frame. To ensure the free-form descriptions, we organize ten persons with different genders and ages to the label language without any limitation of expression styles. To guarantee the correctness and uniqueness of the linguistic description for target objects, we implement a triple-check mechanism. For each description of one annotator, \textit{only the one that three other annotators all correctly localize the target in the scene can remain}. 

\subsection{Statistics}

We uniformly sampled 662 scenes from original STCrowd dataset, including a total length of 65 minutes, for \textbf{STRefer} and annotate 5,458 natural language descriptions for 3,581 subjects. The scene here means a frame of synchronized LiDAR point cloud and image. The content in each scene distinguishes from others due to changing capture locations or time. We split it into training and testing data by 4:1 without data leakage. \textbf{LifeRefer} involves 25,380 natural language descriptions for 11,864 subjects based on 3,172 scenes, which has totally 103 minutes length. Similarly, we split it into 14,650 training data and 10,730 testing data without data leakage.
For both datasets, we ensure each word in the test set appears at least twice in the training set to avoid the influence of data randomness on the experiment. Figure.~\ref{fig:word_cloud} shows commonly used attributes in our descriptions. We also provide the consecutive neighboring frames of raw data for each scene to benefit the leverage of dynamic information. In particular, all the faces in images are masked for \textbf{privacy-preserving}.

\subsection{Characteristics}
Compared with existing open-source datasets for 3D visual grounding, STRefer and LifeRefer have four unique characteristics, listed in Table.~\ref{table:dataset_comparison}. Detailed analyses are as follows. 

\textbf{Human-centric Scenarios. }
Scene understanding for static indoor scenarios and traffic scenarios have made great progress in recent years, which truly boosts the development of autonomous driving and assistive robots. Human-centric scene understanding progresses relatively slowly due to the lack of large-scale open datasets and the intrinsic difficulties of the problem, where humans have abundant poses, actions, and interaction with other humans or objects. Our datasets focus on the challenging human-centric scenarios, which is significant for developing fine-grained interactive robots.

\textbf{Large-scale Dynamic Scenes with Time Series. }
We are living in a dynamic world with moving and changing objects and environments. Our datasets concern the real large-scale dynamic scenes and capture consecutive visual data. When annotating the language for targets, we also pay attention to the description for temporal states, such as short-term or long-term actions. The sequential visual data and time series dependent linguistic descriptions is a distinguishing feature of our datasets.  

\textbf{Multi-modal Visual Data. }
RGB-D camera can only work in indoor scenarios with 5m perception range. To capture the long-range scene data, the popular configuration of visual sensor is the calibrated LiDAR and camera, which provide 3D point cloud and 2D images, respectively. We follow this setting in our dataset to facilitate the downstream robotic applications. However, compared with RGB-D data, the LiDAR point cloud is unordered and much sparser and there are no accurate one-to-one physical correspondence between it and images. Therefore, the fusion strategy of unordered-sparse represented point cloud and regular-dense represented image is not straightforward, which all bring challenges to accurate object grounding.

\textbf{View-dependent Visual Data. }
For the aim of real-time grounding in the wild, we use online captured visual data, which is different from the pre-scanned point cloud of whole scenes~\cite{scanrefer, referit3d}. It is similar to human's observation, where external and self-occlusions exist in crowded scenarios. The benefit is that all spatial relations in our descriptions are all based on the sensor view, which is much clearer than the definition in the global view and is more suitable for application deployment.

\begin{figure*}[t]
    \centering
    \includegraphics[width=1\textwidth]{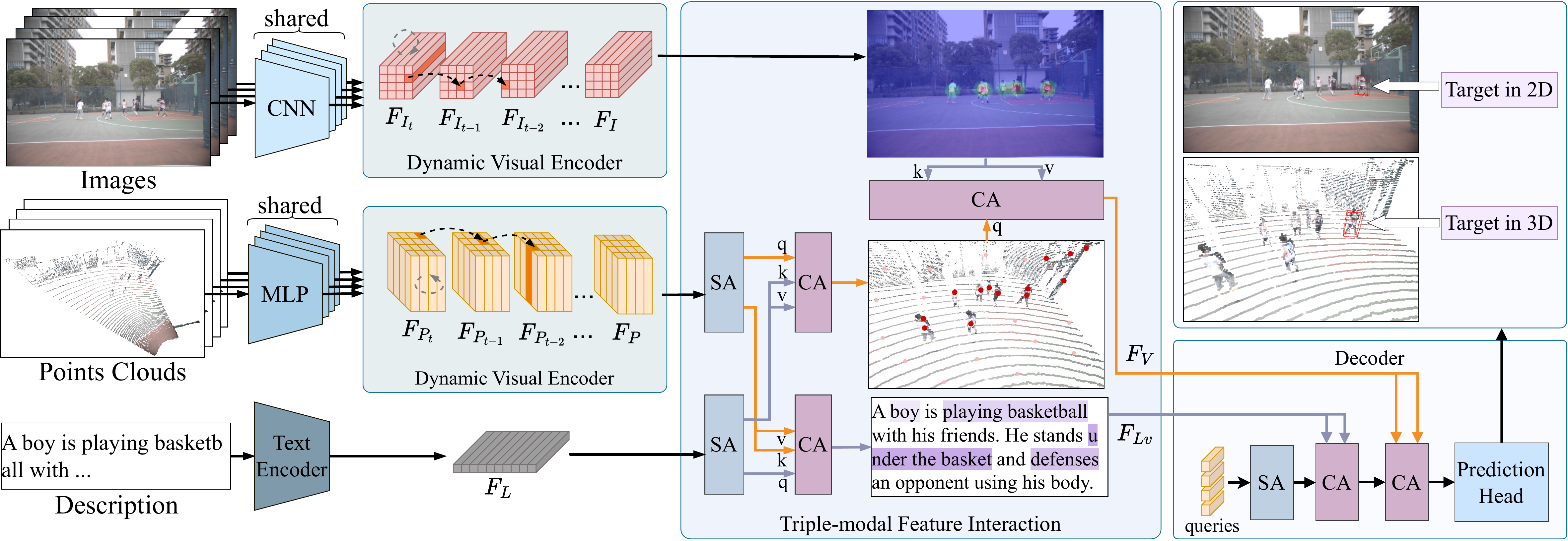} 
    \caption{Pipeline of WildRefer. The inputs are multi-frame synchronized points clouds and images as well as a natural language description. After feature extraction, we obtain two types of visual feature and a text feature. Through two dynamic visual encoders, we extract dynamic-enhanced image and point features. Then, a triple-modal feature interaction module is followed to fuse valuable information from different modalities. Finally, through a DETR-like decoder, we decode the location and size of the target object. SA and CA denote self-attention and cross-attention, respectively.}
    \label{fig:pipeline}
\end{figure*}

\section{Methodology}
\label{sec:method}

\subsection{Overview}
For our 3DVGW task, the inputs include synchronized LiDAR points clouds and images as well as one free-form linguistic description. The output is the 3D bounding box of the target object in physical world. To fully exploit the temporal information existing in sequential data and match with action-related descriptions, we design a useful dynamic visual feature encoder. Moreover, in order to make both the geometry information in point clouds and appearance pattern in images better align with the semantic feature in language, we propose a triple-modal feature interaction strategy for feature enhancement. As shown in Figure.~\ref{fig:pipeline}, our method, WildRefer, solves the 3DVGW problem in the one-stage fashion, including initial feature extractors, a dynamic visual encoder (DVE), a triple-modal feature interaction module (TFI), and a decoder. We introduce details of our method in the following.

\subsection{Feature extraction}
We use Pointnet++~\cite{qi2017pointnet++} to extract the geometry features $\{F_{P_t}, F_{P_{t-1}}, ..., F_{P_{t-K+1}}\}$ from temporal point clouds and pretrained ResNet34~\cite{resnet} to extract appearance features $\{F_{I_t}, F_{I_{t-1}}, ..., F_{I_{t-K+1}}\}$ from temporal images, where $t$ denotes current time, and $K$ denotes the length of input sequence. For utterances, we leverage pretrained RoBERTa~\cite{roberta} to extract text feature $F_L$.

\begin{figure*}[b]
    \centering
    \includegraphics[width=0.6\linewidth]{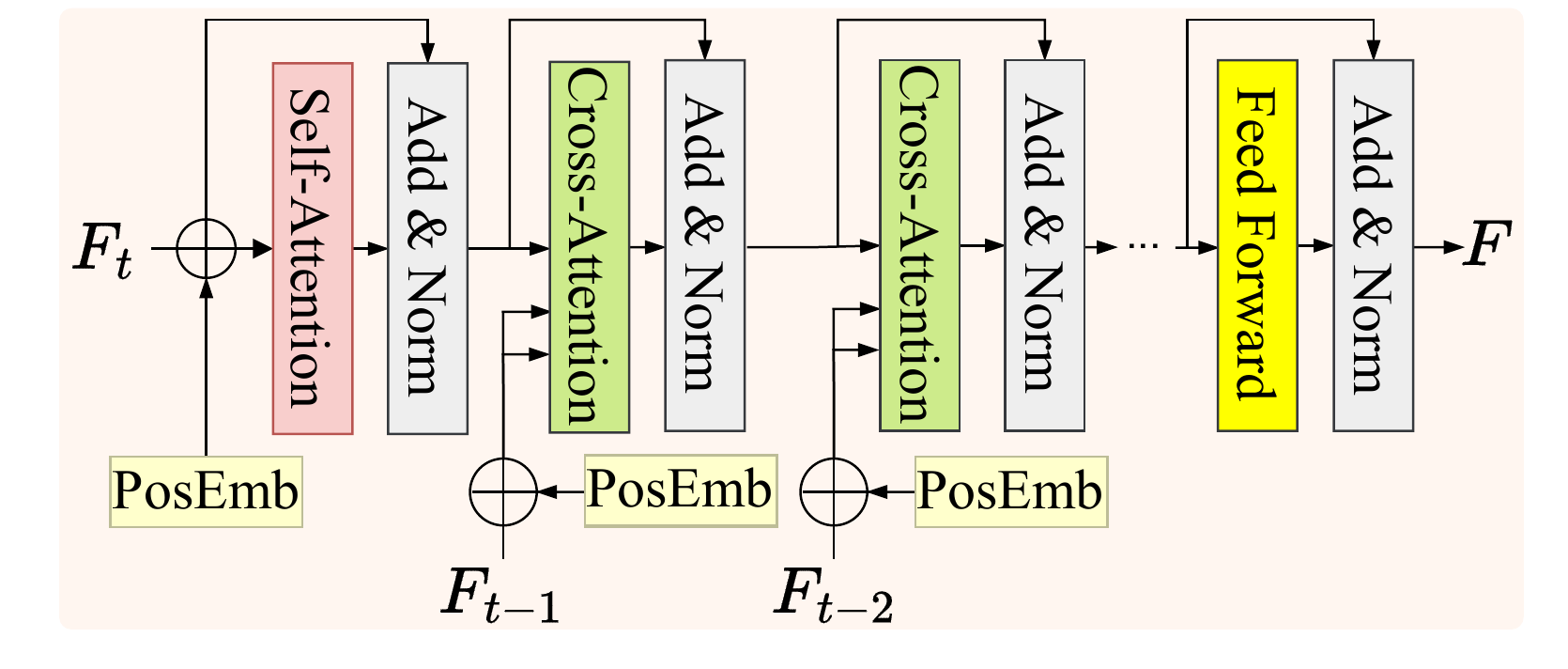}
    \caption{Structure of dynamic visual encoder. $F$ indicates the feature extracted from either point clouds or images, $t$ is the time, and PosEmb means position embedding.}
    \label{fig:dve}
\end{figure*}

\subsection{Dynamic Visual Encoder (DVE)}
Apart from location or texture related descriptions, dynamic motion-related descriptions are also important signals to locate specific human target in large-scale scenes, such as walking, standing, sitting, riding, waving, etc. Moreover, the occlusion problem in dense crowd can also be alleviated by using temporal information. Such dynamic features can be extracted from consecutive visual data, where the changes of depth values in points clouds and pixel values in images represent motion patterns of human actions. Because the scene content is always changing, it is not easy to match the object features from different times in large-scale scenes. Therefore, we apply a spatial-temporal attention mechanism to track critical features of objects automatically. As Figure.~\ref{fig:dve} shows, we first enhance the spatial feature of current time by self-attention. Then, by gradually matching the feature in previous time series through cross-attention, we obtain the dynamic feature $F$ of object. The process of using two adjacent frames is
\begin{equation}
\begin{aligned}
    & ATT(q, k, v) = softmax(\frac{q\cdot k }{\sqrt{C}}) \cdot v, \\
    & F^{'} = ATT(F_t, F_{t-1} + P_{t-1}, F_{t-1} + P_{t-1}), \\
    & F = LN(FFN(F^{'}) + F{'}), \\
\end{aligned}
\label{eq:dve}
\end{equation}
where $P_{t-1}$ denotes the position embedding at $t-1$ time, $LN$ is layer normalization, $FFN$ means feed forward layer.


\subsection{Triple-modal Feature Interaction (TFI)}
At current time, we already have one semantic feature $F_L$ from language and two dynamic visual features $F_P$ and $F_I$ from point clouds and images. We make these triple-modal features cross-attend with each other to benefit the vision-language matching and target localization.

For the visual grounding task in 3D space, the visual feature of point cloud contains rich geometry, position, and motion information, which are critical to match with language descriptions. Moreover, the 3D coordinate-related representation of point feature is also the key for the final target localization. Therefore, we first make the point feature interact with the language feature via a sequence of cross-attention layers, like Figure.~\ref{fig:pipeline} shows. The process is represented by Formula.~\ref{eq:tfi}. After this, we obtain visual-enhanced semantic feature $F_{Lv}$ and the semantic-enhanced point feature $F_{Pl}$.
\begin{equation}
\begin{aligned}
    & F_{Lv} = LN(FFN(ATT(F_L, F_P, F_P) + F_L). \\
     & F_{Pl} = LN(FFN(ATT(F_P, F_L, F_L) + F_P). \\
\end{aligned}
\label{eq:tfi}
\end{equation}

Then, we further enhance the point feature through fusing corresponding dense texture features from images. Because there are no strict one-to-one physical correspondence between 2D and 3D features, we make use of cross-attention to learn effective fusion strategy automatically. Finally, the comprehensive visual feature $F_V$ is calculated by 
\begin{equation}
\begin{aligned}
    & F_V = ATT(F_{Pl}, F_I, F_I). 
\end{aligned}
\label{eq:vles}
\end{equation}

\subsection{Decoder}
Similar to BUTD-DETR\cite{butd-detr}, we use non-parametric queries, selected by $N$ highest confidences $s$. The confidence score is obtained by $F_V$ through a MLP. Then, through cross attention, the queries gradually extract key information from semantic feature and visual feature. Next, the prediction head decodes the location and size of the objects, and a $N\times M$ probability mapping of each visual token and each word in the utterance, where $M$ is the length of the sentence. Finally, we select the bounding box whose corresponding visual feature is furthest with word feature of ``not-mentioned'' as the target location.

\subsection{Training Details}
We use five distinct losses to supervise our training process. Firstly, we use focal loss $L_s$ to supervise the target confidence $s$ to increase the distribution difference between target points and non-target points. Then, we use GIoU loss $L_{giou}$ and L1 loss $L_{box}$ to supervise the position and size of predicted boxes. Following ~\cite{butd-detr}, we use a contrastive loss $L_c$ to supervise the similarity of visual tokens and their corresponding word feature. The visual feature of the target should be similar with the semantic feature of target word in the utterance and different with other words. The structure of sentence is parsed by SpaCy. In addition, one span ``not-mentioned'' is appended at the end of utterances to match non-target objects queries. At last, we use soft token prediction loss~\cite{mdetr} $L_{st}$ to supervise the probability mapping of each visual tokens and words. Finally, taking $\alpha$, $\beta$, $\gamma$, $\lambda$ and $\mu$ as weights of losses, our loss $L$ is calculated by
\begin{equation}
    L = \alpha L_s + \beta L_{giou} + \gamma L_{box} + \lambda L_c + \mu L_{st},
\label{eq:loss}
\end{equation}



\section{Experiments}
\label{sec:experiments}

In this section, we first provide implementation details, and then show extensive experiments, including the comparison with current SOTA methods and ablation studies. More detailed experiments are provided for our feature fusion strategy and the temporal information exploration. We use the accuracy with 3D IOU threshold and mean IOU (mIOU) as metrics and choose 0.25 and 0.5 as IOU thresholds. 

\begin{table}[b]
\centering
\caption{3D Detection results on STRefer and LifeRefer. We show precision (P) and recall (R) with 25\% and 50\% IOU. }
    \small\begin{tabular}{l|cccc}
    \hline
              & R@0.25 & R@0.5  & P@0.25 & P@0.5 \\
    \hline
    STRefer   & 90.77   & 82.13 & 91.69 & 83.01 \\
    LifeRefer & 77.11   & 43.16 & 60.09 & 33.63 \\
    \hline
    \end{tabular}
\label{table:detection}
\end{table}

\subsection{Implementation Details}
We use ResNet34~\cite{resnet} and PointNet++~\cite{qi2017pointnet++} to extract visual features, and RoBERTa~\cite{roberta} to extract language features. All features are projected to 288-D. 
We use 1-layer and 6-layer transformers in the DVE and decoder, respectively. For the cross-attention in TFI, we set 3 layers. We extract dynamic features from consecutive two frames (spanning one second) of data. For all layers in the transformer, we use 8-head attention and 256-D feed-forward layer with 10\% dropout. 
We train 200 epochs on STRefer and 60 epochs on LifeRefer by AdamW~\cite{adamw} with 0.0001 learning rate and 0.0005 weight decay. For PointNet++\cite{qi2017pointnet++} and RoBERTa~\cite{roberta}, the initial learning rates are 0.001 and 0.00001. The learning rate decays by a factor of 10 after 45 epochs on LifeRefer. The weights of losses $\alpha$, $\beta$, $\gamma$, $\lambda$ and $\mu$ are set as 8, 1, 5, 1 and 0.01.

\subsection{Comparison}

\begin{table}[t]
\centering
\caption{Comparison results on STRefer and LifeRefer. {*} denotes the one-stage version without pretrained 3D decoder.}
\resizebox{\linewidth}{!}{
\begin{tabular}{l|c|c|ccc|ccc|c}
    \hline
    Method   &Publication    & Type & \multicolumn{3}{c|}{STRefer} & \multicolumn{3}{c|}{LifeRefer}  & Time cost   \\
                &  &     & Acc@0.25  & Acc@0.5  & mIOU  & Acc@0.25  & Acc@0.5  & mIOU    & (ms)                      \\
    \hline
    ScanRefer~\cite{scanrefer}  & ECCV-2020        & Two-Stage  & 32.93   & 30.39  & 25.21    & 22.76 & 14.89     & 12.61      & 156    \\
    ReferIt3D~\cite{referit3d}  & ECCV-2020        & Two-Stage  & 34.05   & 31.61  & 26.05    & 26.18 & 17.06     & 14.38      & 194    \\
    3DVG-Transformer~\cite{3dvgtransformer} & ICCV-2021  & Two-Stage  & 40.53   & 37.71  & 30.63    & 25.71 &  15.94    & 13.99 & 160    \\
    MVT~\cite{mvt}     &  CVPR-2022   & Two-Stage  & 45.12   & 42.40  & 35.03    & 21.65 &  13.17    & 11.71 & 242    \\
    3DJCG~\cite{3djcg}  &  CVPR-2022  & Two-Stage  & 50.47   & 47.47  & \bf{38.85}    & 27.82 &  16.87    & 15.40 & 161    \\
    BUTD-DETR~\cite{butd-detr}  & ECCV-2022 & Two-Stage  & 57.60   & 47.47  & 35.22    & 30.81 &  11.66    & 14.80 & 252    \\
    EDA~\cite{eda}       & CVPR-2023  & Two-Stage  & 55.91   & 47.28  & 34.32    & 31.44 &  11.18    & 15.00 &  291      \\
    \hline
     3D-SPS~\cite{3dsps}  &   CVPR-2022  & One-Stage  & 44.47   & 42.40  & 30.43    & 28.01 &  18.20    & 15.78 & 130    \\
    BUTD-DETR$^{*}$~\cite{butd-detr}  & ECCV-2022  & One-Stage  & 56.66   & 45.12  & 33.52    & 32.46 &  12.82    & 15.96 &  138    \\
    EDA$^{*}$~\cite{eda}       & CVPR-2023  & One-Stage  & 57.41   & 45.59  & 34.03    & 29.32 &  12.25    & 14.41 &  154      \\
    \hline
    Ours      &--        & One-Stage  & \bf{62.01}   & \bf{54.97} & 38.77   & \bf{38.89} &  \bf{18.42}    & \bf{19.47} & 151       \\
    \hline
    \end{tabular}
    }
\label{table:method_comp}
\end{table}

Because we are the first to solve 3DVGW task based on multi-sensor visual data in dynamic scenes, there is no related work for absolutely fair comparison. 
In order to verify the effectiveness of our method, we compare with recent SOTA methods for 3D visual grounding based on RGB-D scans, including ScanRefer~\cite{scanrefer}, ReferIt3D~\cite{referit3d}, 3DVG-Transformer~\cite{3dvgtransformer}, Multi-View Transformer (MVT)~\cite{mvt}, 3DJCG~\cite{3djcg}, 3D-SPS~\cite{3dsps}, BUTD-DETR~\cite{butd-detr} and EDA~\cite{eda} as Table.~\ref{table:method_comp} shows. To achieve fair comparisons as much as possible, we project the LiDAR point cloud on the image by calibration metrics and paint points with the corresponding color on the image to obtain the pseudo RGB-D data. We conduct experiments by using their released code. For two-stage methods, we use current state-of-the-art 3D detector~\cite{stcrowd} pretrained on STCrowd and LifeRefer to provide 3D proposals in the first stage. The detection performance is shown in Table.~\ref{table:detection}. Because BUTD-DETR and EDA can run without pre-detection inputs, we also run their one-stage versions for the comparison. For datasets, because KITTI360Pose~\cite{text2pos} focuses on the localization, their grounding target is not objects with bounding boxes but only global positions, which is greatly different from our task. Thus, we cannot conduct experiments on it.

\begin{figure*}[t]
  \centering
    \includegraphics[width=1\linewidth]{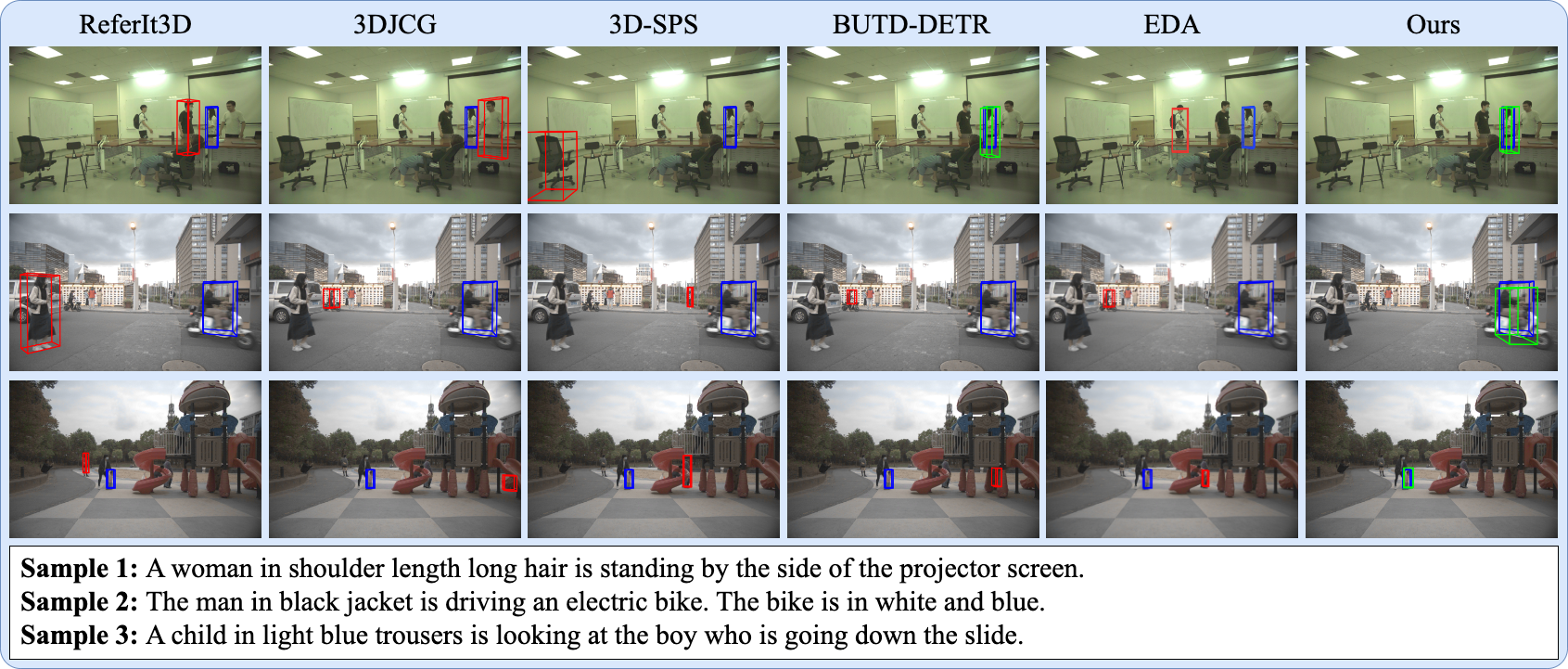}
  \caption{Visualization of comparison results on LifeRefer. Blue, red, and green boxes denote the ground truth, wrong prediction, and right prediction, respectively.}
  \label{fig:comparison}
\end{figure*}

It can be seen from the results that our method outperforms others obviously. Previous 3D visual grounding methods rely on RGB-D representation where points and colors have physical dense correspondence. While for the LiDAR and camera setting, direct painting colors on sparse LiDAR point cloud will miss lots of texture features in images. Our method utilizes cross-attention to learn valuable features from different modalities for fusion, keeping more critical visual information for the further vision-language matching. Furthermore, our one-stage framework can get rid of the dependence on 3D detector. From Table.~\ref{table:detection}, we can see that scenarios in LifeRefer is much more complex than that of STRefer, causing low detection accuracy. An interesting phenomenon is that the one-stage version of BUTD-DETR is superior than its two-stage version due to the low-quality detected proposals. Although two-stage methods have more prior knowledge by pretraining with more data for supervision, one-stage framework still achieves better performance especially for in-the-wild complex scenarios. Moreover, our method extracts dynamic features of real scenes, which can better match motion-related descriptions for more accurate localization. In addition, we also show the time cost of inference in Table.~\ref{table:method_comp}. Although our method is required to process multi-frame data and using transformer mechanism for feature interaction, it is still efficient and has the potential to be applied in real-time applications. The qualitative comparison is illustrated in Figure.~\ref{fig:comparison}.

\begin{figure}[h]
  \centering
  \includegraphics[width=0.8\linewidth]{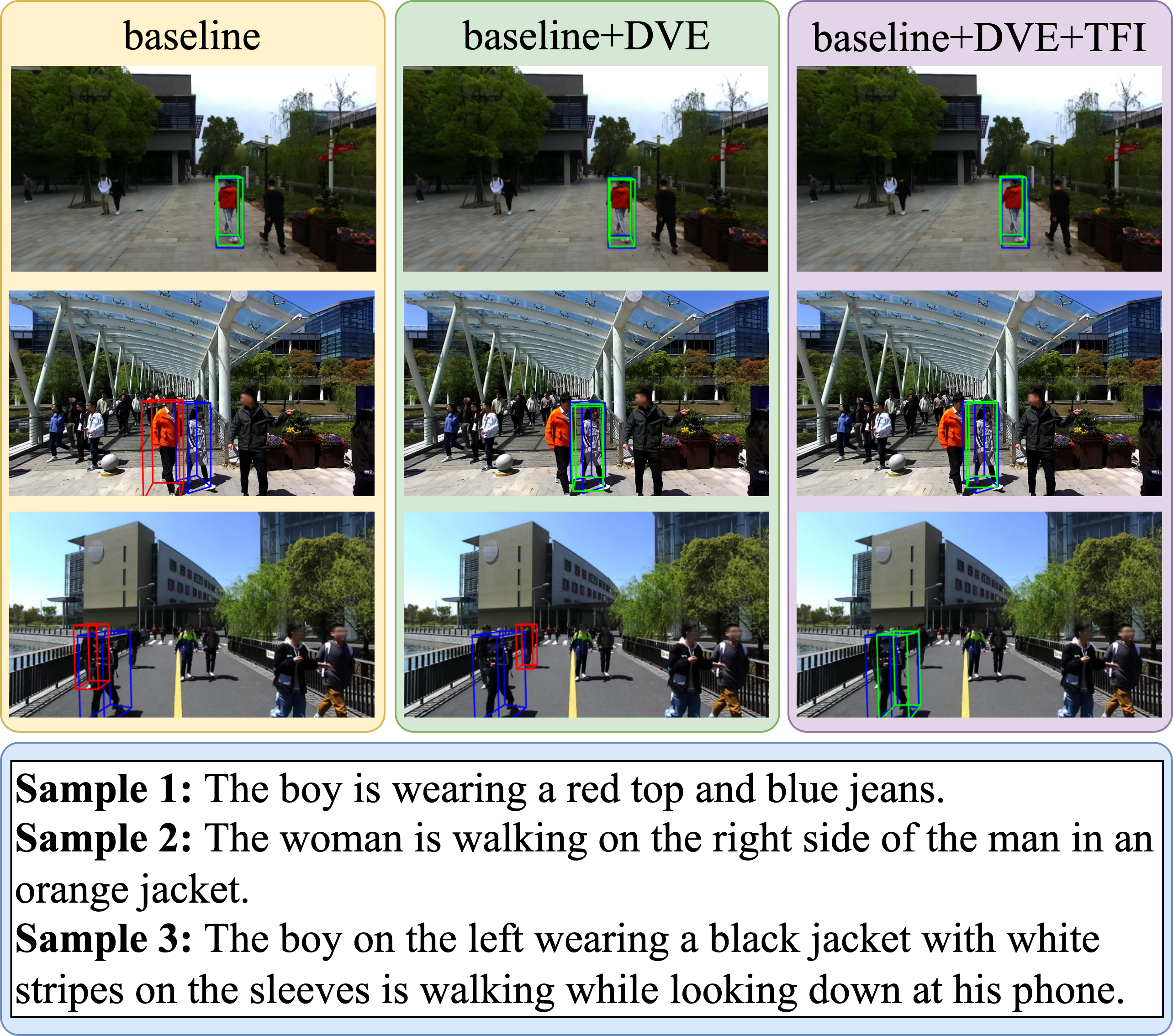}
  \caption{Visualization results of ablation results on STRefer.}
  \label{fig:ablation}
\end{figure}

\subsection{Ablation Studies}

Quantitative and qualitative ablation results are demonstrated in Table.~\ref{table:ablation} and Figure.~\ref{fig:ablation}. As Figure.~\ref{fig:ablation} shows, the detailed feature ``white stripes'' cannot be recognized by point cloud-only representation, but can be recognized by fusing two modal visual features.

\noindent \textbf{Baseline.} 
For the baseline network, we only use one frame of point cloud painted by color as the visual input and only conduct point-language interaction by cross-attention, which is same with BUTD-DETR. 

\noindent \textbf{Baseline + DVE.}
The Dynamic Visual Encoder (DVE) is attached after point feature extractor in baseline network to make full use of dynamic information. After spatial-temporal attention, we obtain the motion feature of objects, which is helpful to localize the target by matching the dynamic information in descriptions. 

\noindent \textbf{Baseline + DVE + TFI.} 
This is the complete network of our method. Triple-model feature interaction module (TFI) leverage dense appearance features in images to compensate for the shortcomings of sparse point representation.

\begin{table}[t]
\centering
\tabcolsep=0.12cm
\caption{Ablation studies of our method on STRefer and LifeRefer.}
\small\begin{tabular}{c|lccc}
    \hline
     &       & Acc@0.25    & Acc@0.5     & mIOU    \\
    \hline
    & baseline        &  56.66  &  45.12  &  33.52 \\
    STRefer & + DVE    &  59.76  &  50.28  &  36.59 \\
    & + DVE + TFI    &  62.01  &  54.97  &  38.77   \\
     \hline
            & baseline & 32.46 &  12.82    & 15.96 \\
    LifeRefer & + DVE          &  38.27    &  16.42    & 18.83  \\
             & + DVE + TFI  & 38.89   & 18.42   & 19.47   \\
    \hline
\end{tabular}
\label{table:ablation}
\end{table}

\subsection{Multi-modal Fusion Strategy}
In our setting, we have three modalities and related features, including the language, point clouds, and images. For locating the target, it is necessary to obtain valuable information for the object from all modal data and conduct accurate mapping from visual feature to linguistic feature. For our triple-modal feature interaction module, we make point feature and language feature interact with each other first and then fuse image features with the semantic guidance. We compare with other feature-fusion orders in Table.~\ref{table:order}. 
For the setting of Vision First, we firstly fuse image feature into points feature and then interact with language feature. It can obtain a comparable result but a bit lower than ours, mainly because the fused visual feature is more complicated for language feature to find corresponding information.
For the setting of Image Domin., we use image as main feature to cross-fuse with language feature first and as queries to extract point clouds feature. The performance drops on STRefer, mainly because we aim to locate targets in 3D space and a point-dominated feature with accurate position information is more suitable than image-dominated feature.

\subsection{Temporal Information Exploration}
Table \ref{table:dve_ablation} shows the influence of frame numbers on DVE and other temporal fusion strategies. Considering both efficiency and accuracy, we choose 2 frames in our setting.
Furthermore, to further evaluate DVE, we compare with other popular dynamic fusion strategies shown in the last two rows. In. Con. denotes directly splicing point clouds of two frames and Fe. Con. means aggregating points and concatenating features of two frames. Our method outperforms others, showing the superiority of our DVE.

\begin{table}[t]
\centering
{
\fontsize{7pt}{10pt}\selectfont
\begin{minipage}{.48\textwidth}
    \centering
    \tabcolsep=0.05cm
    \caption{Comparison of other visual fusion strategies.}
    \begin{tabular}{c|lccc}
        \hline
         &       & Acc@0.25    & Acc@0.5     & mIOU    \\
        \hline
                & Vision First        & 59.74  & 52.72   & 37.11 \\
        STRefer & Image Domin.    & 52.25  & 47.47   & 32.80 \\
                & ours &  62.01  &  54.97  &  38.77   \\
                  \hline
                & Vision First & 37.27  & 16.92   & 18.65 \\
        LifeRefer & Image Domin.   & 38.82  & 16.78   & 18.87  \\
                 & ours  & 38.82   & 18.42   & 19.47 \\
        \hline
    \end{tabular}
    \label{table:order}
\end{minipage}
}
\hfill
{
\fontsize{7pt}{10pt}\selectfont
\begin{minipage}{.48\textwidth}
    \centering
    \caption{More experiments of DVE on STRefer.}
    \begin{tabular}{c|c|ccc}
        \hline
        \multicolumn{1}{l|}{Strategy} & Frames  & Acc@0.25  & Acc@0.5  & mIOU    \\
        \hline
        \multirow{4}{*}{DVE} & 1 &  59.38  &  51.69  & 36.99  \\
        & 2 &  \textbf{62.01}  &  \textbf{54.97}  & \textbf{38.77}  \\
        & 3 &  59.94  &  54.60  & 37.76  \\
        & 4 &  60.41  &  52.25  & 36.89  \\
        \hline
        In. Con. & 2 & 60.79   & 53.19   & 37.88  \\
        Fe. Con. & 2 & 40.24   & 29.64   & 23.14  \\
        \hline
    \end{tabular}
    \label{table:dve_ablation}
\end{minipage}
}
\end{table}



\section{Conclusion}
\label{sec:conclusion}

In this paper, we first introduce the task of 3DVGW, which focuses on the large-scale dynamic scenes and uses online captured multi-modal visual data. Then, we propose effective vision-language interaction and sensor-fusion strategy for 3D vision grounding by fully utilizing the geometry feature, appearance feature, and motion feature. In particular, we propose two novel datasets for 3DVGW on human-centric scenarios and hope our work could inspire more related research for developing AI robots. In the future, it is promising to extend them with more comprehensive descriptions for more kinds of objects in scenes.


\bibliographystyle{splncs04}
\bibliography{main}
\end{document}


\title{WildRefer: 3D Object Localization in Large-scale Dynamic Scenes with Multi-modal Visual Data and Natural Language} 

\titlerunning{WildRefer}

\author{Zhenxiang Lin\inst{1} 
\and
Xidong Peng\inst{1} 
\and
Peishan Cong\inst{1}
\and
Ge Zheng\inst{1}
\and
Yujin Sun\inst{2}
\and
Yuenan Hou\inst{3}
\and
Xinge Zhu\inst{4}
\and
Sibei Yang\inst{1}
\and
Yuexin Ma\inst{1}\thanks{Corresponding author. This work was supported by NSFC (No.62206173), Natural Science Foundation of Shanghai (No.22dz1201900), Shanghai Sailing Program (No.22YF1428700), MoE Key Laboratory of Intelligent Perception and Human-Machine Collaboration (ShanghaiTech University), Shanghai Frontiers Science Center of Human-centered Artificial Intelligence (ShangHAI).}
}

\authorrunning{Lin et al.}

\institute{
$^1$ShanghaiTech University 
$^2$The University of Hong Kong \\
$^3$Shanghai AI Laboratory 
$^4$The Chinese University of Hong Kong \\
\email{zhenxianglin94@gmail.com}, 
\email{mayuexin@shanghaitech.edu.cn}
}

\maketitle

\appendix

In the appendix, we provide more details of STRefer and LifeRefer dataset and more experimental results. Firstly, we show some visualization examples of our datasets, and provide some samples to validate the effectiveness of our method as well as visualization cases of method comparison. Furthermore, we show the formula of evluation metrics, and provide some additional experiments. Then, we discuss the feasibility of 2D video grounding methods with 3D projection. At last, we show the limitation of our method and analyze the reason and potential solutions. Our code has been packaged and uploaded as supplementary material for proving that our work is reproducible and all the experimental results are convincing. The code is provided in \url{https://github.com/4DVLab/WildRefer}.

\section{Dataset Visualization}
\begin{figure}[h]
  \centering
  \includegraphics[width=1\linewidth]{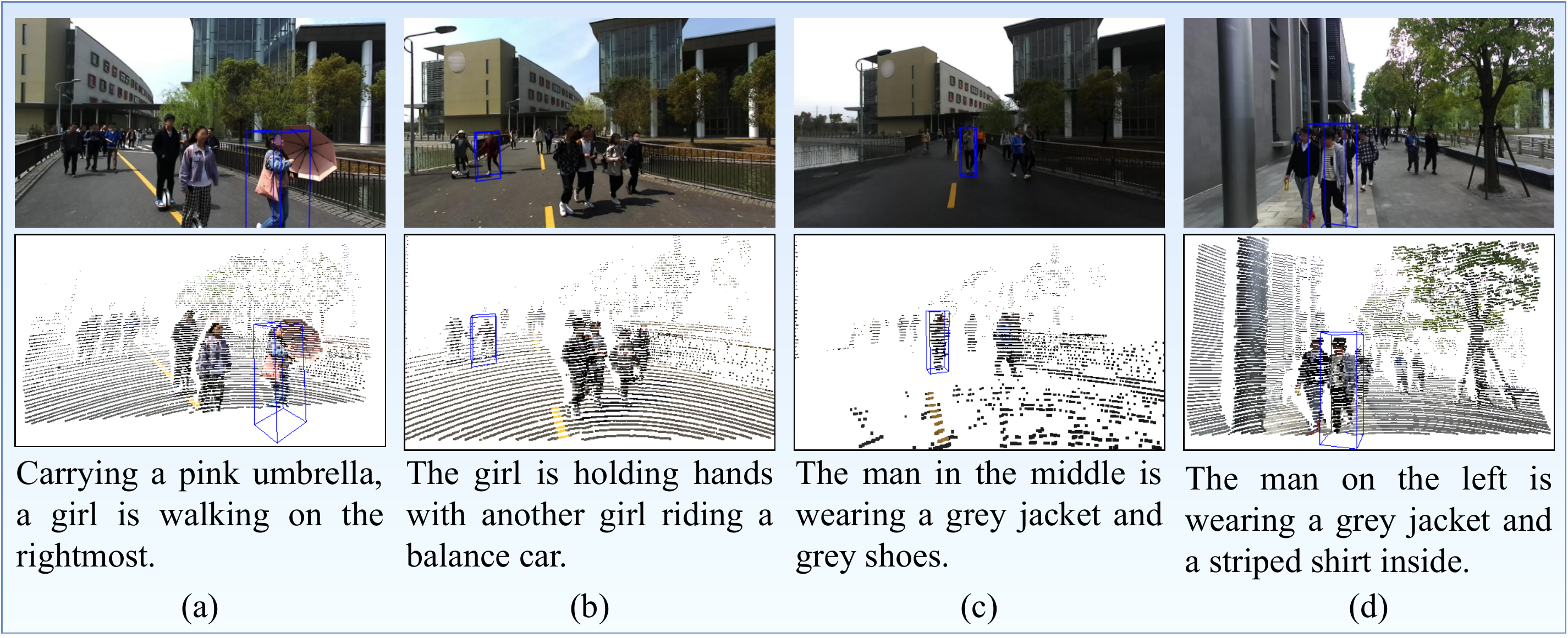}
  \caption{Visualization of STRefer dataset. Blue boxes are the targets.}
  \label{fig:dataset_visualization_strefer}
\end{figure}

\begin{figure}[h]
  \centering
  \includegraphics[width=1\linewidth]{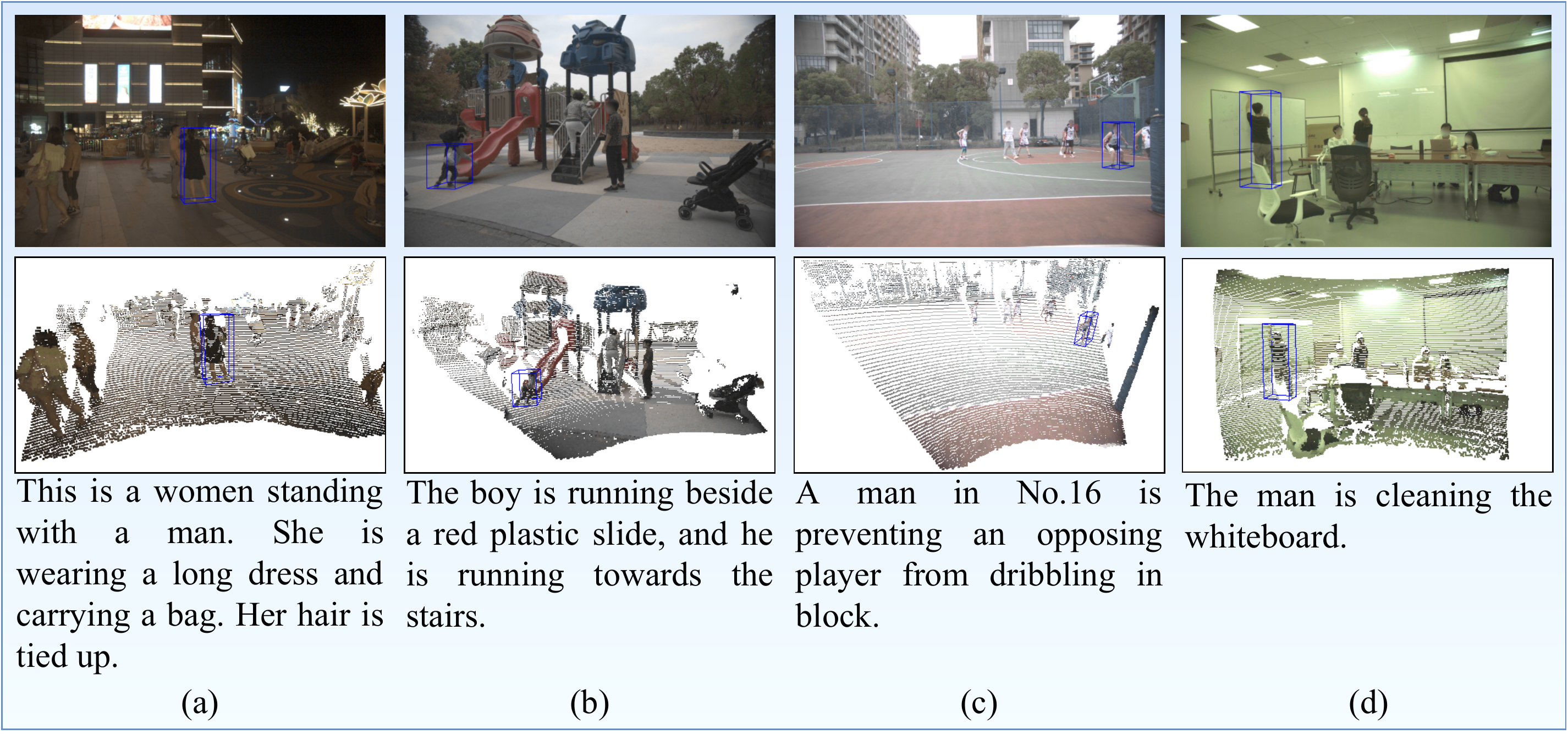}
  \caption{Visualization of LifeRefer dataset. Blue boxes are the targets.}
  \label{fig:dataset_visualization_liferefer}
\end{figure}

In this section, we provide visualization examples of our dataset for more intuitive understanding for STRefer and LifeRefer. As can be seen from Figure.~\ref{fig:dataset_visualization_strefer} and Figure.~\ref{fig:dataset_visualization_liferefer}, the target scenes are crowds with various amounts of pedestrians and different backgrounds. Each tuple of data consists of the image (from camera), point cloud (from LiDAR) and the corresponding human-written free-form description. We can locate the target by pose (walking, riding, standing, cleaning, etc.), color (pink, grey, red, etc.), spatial relation (left, right, etc.), some objects (umbrella, balance car, slide, whiteboard, etc.) or some other attributes.

\section{Results Visualization}
\begin{figure}[h]
  \centering
  \includegraphics[width=1\linewidth]{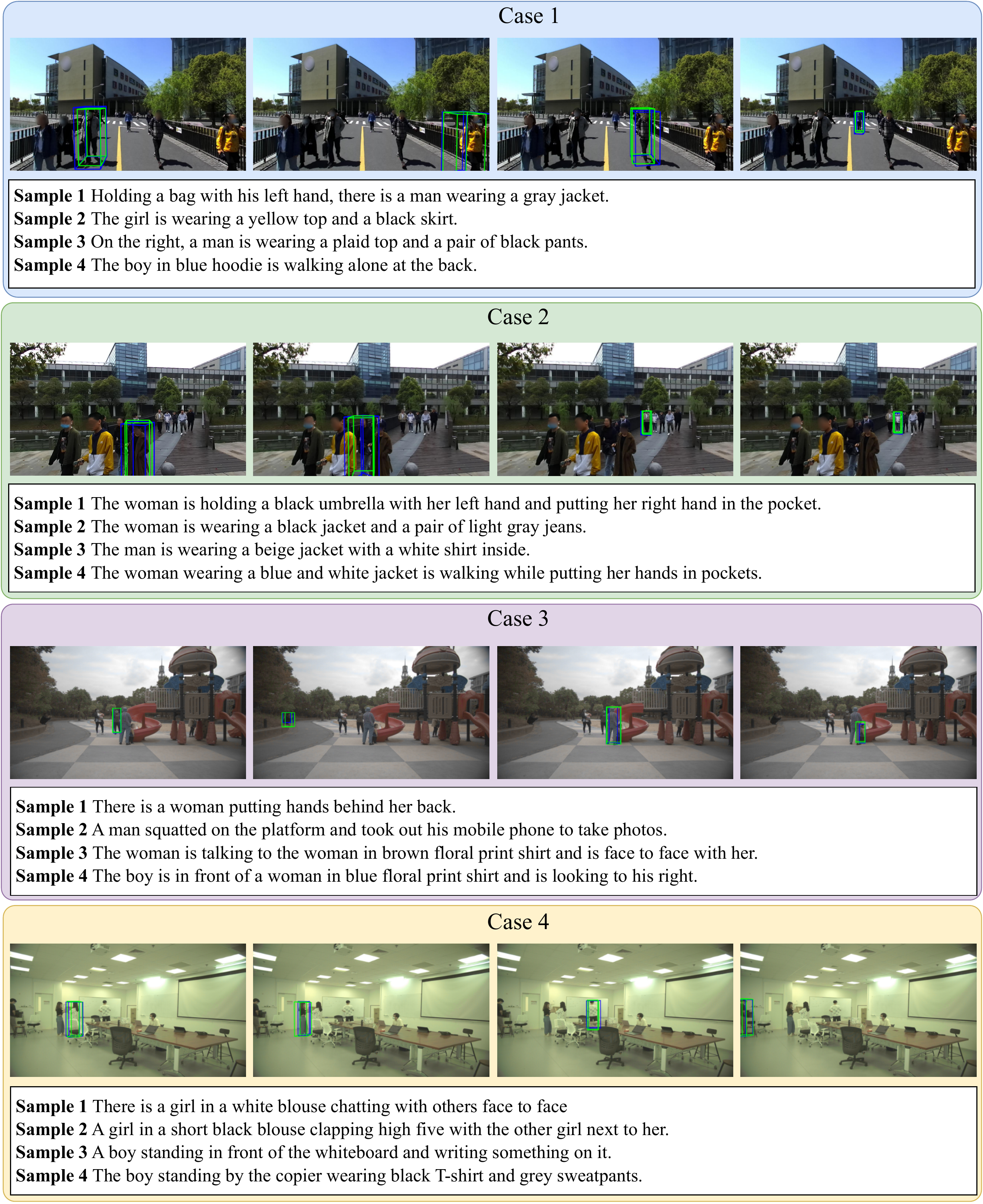}
  \caption{Results of our method. Blue and green boxes denote the ground truth and our prediction results, respectively. Case 1 and 2 are on STRefer. Case 3 and 4 are on LifeRefer.}
  \label{fig:method_show}
\end{figure}

In order to verify the effectiveness of our method, we analyze the results which are shown in Figure.~\ref{fig:method_show}. As can be seen, our method can locate the target by different descriptions and can distinguish different objects by different linguistic information, which indicates our method can extract linguistic feature correctly, and apply it into the 3DVGW task.


\newpage
\section{Evaluation Metrics}
The evaluation metrics are calculated by Formula.~\ref{eq:metrics}, where $|\Omega|$ is the number of samples in dataset.
\begin{equation}
\begin{aligned}
& Acc(IOU_{thr}) = \frac{1}{|\Omega|}\sum_{i=1}^{|\Omega|}I(IOU_{i} \geq IOU_{thr}), \\
& mIOU = \frac{1}{|\Omega| }\sum_{i=1}^{|\Omega|}IOU_{i}.
\end{aligned}
\label{eq:metrics}
\end{equation}

\section{Additional Experiments}
\subsection{Additional Fusion Strategy}
C.F. denotes we directly concatenate two features and fuse them through a MLP, shown in Table~\ref{table:orderAP}. It can be seen that our attention-based mechanism shows the superiority.

\begin{table}[ht]
\centering
\tabcolsep=0.12cm
\caption{Comparison of other visual fusion orders.}
\begin{tabular}{c|lccc}
    \hline
     &       & Acc@0.25    & Acc@0.5     & mIOU    \\
    \hline
     \multirow{2}{*}{STRefer} & C.F.    & 57.22  & 50.28   & 35.13    \\
             & ours &  62.01  &  54.97  &  38.77   \\
              \hline
    \multirow{2}{*}{LifeRefer}   & C.F  & 32.62   & 15.16  & 16.29   \\
             & ours  & 38.88   & 18.43   & 19.47 \\
    \hline
\end{tabular}
\label{table:orderAP}
\end{table}

\subsection{Mean and stand errors comparison between DVE and In. Con.}
We provide the mean and standard errors of four additional experiments with different random seeds in Table~\ref{table:dve_incon}. DVE performs better than In. Con consistently, especially for the most critical metric Acc@0.25. Because DVE can automatically learn the corresponding features from different frames for effective feature fusion in the dynamic scenes. 

\begin{table}[h]
\centering
\caption{Mean and stand errors comparison between DVE and In. Con. on STRefer.}
\tabcolsep=0.12cm
\begin{tabular}{c|c|ccc|ccc|ccc}
\hline
\multirow{2}{*}{Strategy} & \multirow{2}{*}{Frames} & \multicolumn{3}{c|}{Acc@0.25} & \multicolumn{3}{c|}{Acc@0.5} & \multicolumn{3}{c}{mIOU} \\
          &   & best  & mean  & std & best & mean  & std  & best& mean  & std        \\ \hline
DVE       & 2  &\textbf{62.01} & \textbf{59.36} & \textbf{1.55} &\textbf{54.97} & \textbf{52.81} & 1.28 &\textbf{38.77} & \textbf{37.05} & 1.00       \\
In. Con.  & 2  &60.98& 57.90 & 2.34 &53.28 & 51.81 & \textbf{1.00} &37.50 & 36.25 & \textbf{0.98}       \\ \hline
\end{tabular}
\label{table:dve_incon}
\end{table}

\section{Feasibility discussion of 2D video grounding with 3D projection}
\begin{figure}[ht]
  \centering
  \includegraphics[width=1\linewidth]{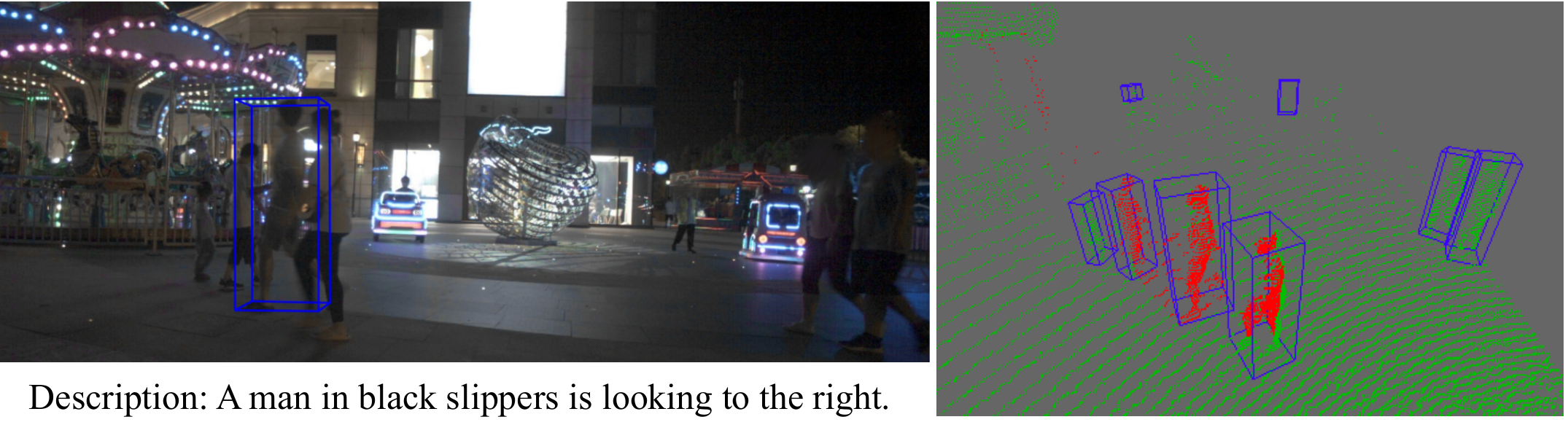}
  \caption{A sample scene of image and LiDAR at night.}
  \label{fig:night}
\end{figure}

2D video grounding methods, designed for video analysis, cannot be used for 3D applications directly. Without accurate depth estimation, the projection from 2D to 3D is ill-posed. If using calibration matrix, many 3D points will correspond to the same pixel position on the image. As Figure~\ref{fig:night} shows, all red points on the right can be projected in the blue bounding box on the left image.
In addition, the video grounding method cannot work in dark scenarios, while LiDAR can work day and night. Our multi-sensor setting can improve the robustness of perception. As Figure~\ref{fig:night} shows, LifeRefer and STRefer contain many data captured at night, it is hard to detect instances from images.

\section{Limitation}
\begin{figure}[ht]
  \centering
  \includegraphics[width=1\linewidth]{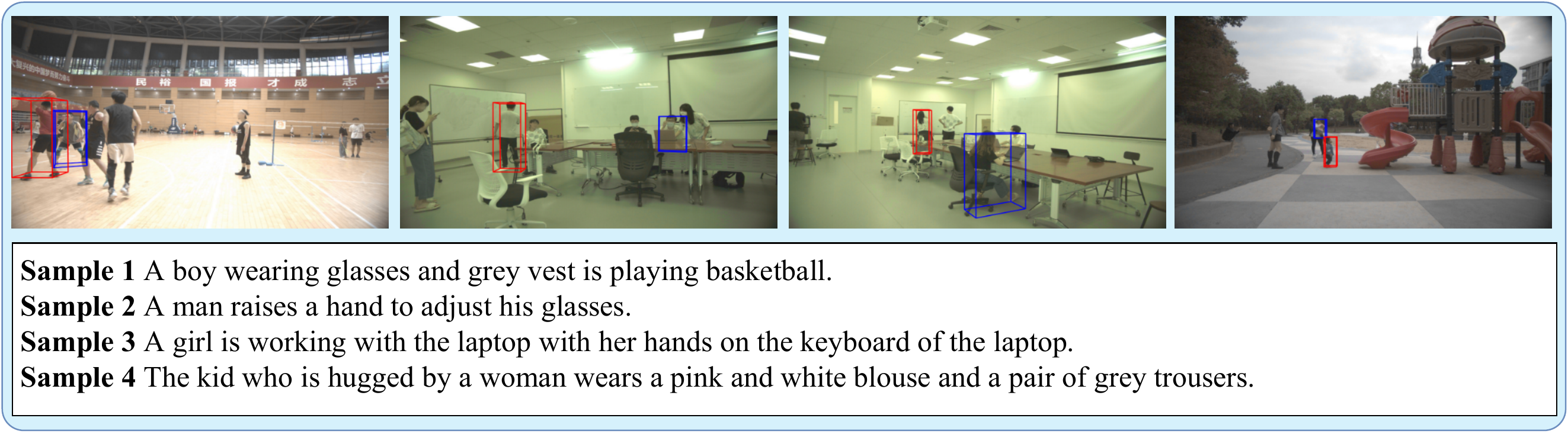}
  \caption{Results of failure cases on LifeRefer. Blue and red boxes denote the ground truth and our prediction results, respectively.}
  \label{fig:method_wrong}
\end{figure}

In this section, we show some failure cases and analyze the potential reason. As can be seen in Figure~\ref{fig:method_wrong}, due to long-tailed distribution, some object words and action words, such as glass, laptop, adjust, etc., appear few times, which cause a difficulty for the learning of the model. In the future, we will expand the scale of the dataset including both the diversity of scenes and descriptions to alleviate the impact of long-tailed problem, and introduce few shot and zero shot methods to increase the generalization of the model.


